\documentclass[
]{ceurart}

\usepackage{amsmath}
\usepackage{amssymb}
\usepackage{enumitem}
\usepackage{comment}
\usepackage{float}
\usepackage{graphicx, color}
\usepackage[utf8]{inputenc}
\usepackage[english]{babel}
\usepackage{multirow}
\usepackage{subfigure}
\usepackage{booktabs}
\usepackage{adjustbox}
\usepackage{hyperref}
\usepackage{pgfgantt}
\usepackage{geometry}
\usepackage{tikz}
\usepackage{subcaption}
\usepackage{tabularx}
\newcolumntype{L}{>{\raggedright\arraybackslash}X} 
\usepackage{arydshln}
\usepackage{placeins}
\usepackage[table,svgnames,dvipsnames,x11names]{xcolor}
\usepackage{listings}
\usetikzlibrary{arrows.meta, positioning}

\AtBeginDocument{%
  }
\begin{document}

\copyrightyear{2026}
\copyrightclause{Copyright for this paper by its authors.
  Use permitted under Creative Commons License Attribution 4.0
  International (CC BY 4.0).}

\conference{[PromptEng] Third International Workshop on Prompt Engineering for Pre-Trained Language Models co-located with the ACM WebConf, April.13, 2026, Dubai, United Arab Emirates}

\title{TAPR: Enhancing LLM Performance with a Task-Aware Prompt Rewriter}

\author[1]{Oliver Savolainen}[%
email=oliver.savolainen@student.uva.nl,
]
\address[1]{University of Amsterdam, Amsterdam, The Netherlands}

\author[2]{Emanuele Bastianelli}[%
email=e.bastianelli@elsevier.com,
]
\address[2]{Elsevier, Amsterdam, The Netherlands}

\author[2]{Hosein Azarbonyad}[%
email=h.azarbonyad@elsevier.com,
]

\author[1]{Ana Lucic}[%
email=a.lucic@uva.nl,
]

\begin{abstract}
Large Language Models (LLMs) often require carefully crafted prompts to unlock their full potential, which can be a barrier for non-expert users. This work addresses the challenge by introducing a Task-Aware Prompt Rewriter (TAPR), a model that reformulates user prompts into task-optimized prompts with the explicit goal of improving downstream LLM performance. We train TAPR using reinforcement learning with Group Relative Policy Optimization (GRPO), where rewards are derived from LLM-as-judge evaluations of both the reformulated prompt and the corresponding task output. Experimental results on diverse tasks, such as question answering, summarization, and arithmetic reasoning, show that our method yields consistent gains over base models in prompt rewriting ability. Fine-tuning Phi-4-mini-instruct (as the base model for TAPR) produces prompts that contain clearer and more instructive language, leading to higher accuracy on established benchmarks such as Natural Questions and GSM8K. Our code is available at: \url{https://github.com/OliverSavolainen/task-specific-prompt-rewriter}
\end{abstract}



\maketitle

\section{Introduction}
\label{sec:intro}

Large Language Models (LLMs) have revolutionized the field of natural language processing, providing unprecedented capabilities in text generation, comprehension, and various other applications \cite{brown2020language, wei2022emergent}. Despite their impressive performance, these models often require carefully crafted input prompts to produce the best possible responses \cite{brown2020language,srivastava2022beyond}. Users, particularly those without expertise in prompt engineering, may find it challenging to formulate prompts that fully leverage the capabilities of these advanced models. This gap underscores the necessity for an automated system that can refine and optimise user prompts, making high-quality interaction with LLMs more accessible \cite{autop, prewrite}. 

Existing methods for prompt optimization rely largely on manual adjustments and iterative testing, which are not only time-consuming but also require a certain level of expertise. Automated prompt rewriting can bridge this gap by refining under‐specified queries into clear, robust instructions that ensure reliable performance for all users. 

In this paper, we present a \textbf{Task-Aware Prompt Rewriter (TAPR)}, a lightweight model that automatically refines and adapts user prompts for LLMs, with the aim of enhancing downstream task performance.
This system involves a small, specialized model dedicated to rewriting prompts, coupled with a larger LLM that executes the task based on these optimized prompts. To train TAPR, we build on the recent advancements on using reinforcement learning to improve LLM performance across a range of tasks.
Therefore, the main research question of this work is:

\noindent\textbf{How can we create a task-aware prompt rewriter such that it enhances the performance of LLMs in downstream tasks?}

We use reinforcement learning to train a smaller LLM to optimize default prompts by rewriting them, using feedback derived from the performance of a frozen Task LLM responding to the rewritten prompts. Prior work has explored using reinforcement learning for this purpose. PRewrite \cite{prewrite} applied PPO (Proximal Policy Optimization) \cite{ppo} to fine-tune LLMs on individual datasets, yielding a single optimal prompt per dataset and demonstrating gains on several benchmarks. 

Our approach extends this in two key ways. First, beyond the verifiable task-performance rewards used by prior work, we introduce using LLM-as-a-judge rewards for semantically more accurate evaluations, including a reward based on the quality of the prompts themselves, and we validate its impact through a dedicated ablation study. Second, we employ Group Relative Policy Optimization (GRPO) \cite{grpo} in place of PPO and achieve great consistency with this algorithm. We also test generating multiple prompts per dataset and having TAPR select the best one, but we do not find this addition leading to further performance gains consistently.

To evaluate our method, we measure the performance of the Task LLM before and after rewriting on various tasks, including question answering, summarization, and arithmetic reasoning. Our framework leads to improvement in the prompt rewriting abilities of Phi-4-mini-instruct and LLaMA-3.2-3B-Instruct on multiple tasks. For Phi-4-mini-instruct, the highest metric score comes from a TAPR variant for each dataset.

\noindent In summary, our contributions are:

\begin{itemize}
  \item We introduce \textbf{TAPR}, a task-aware prompt rewriter trained with reinforcement learning to optimize task-specific prompts using feedback derived from both rewritten prompts and responses from LLM performing the tasks.

  \item We leverage the LLM-as-a-judge paradigm to obtain semantically informed feedback and design a dedicated \textbf{prompt quality reward}, demonstrating that using LLM-based evaluations improves training stability and effectiveness.

  \item We perform various experiments across question answering, summarization, and arithmetic reasoning benchmarks, showing that prompts rewritten by TAPR enhance task performance compared to baseline prompts and rewrites from base models.
\end{itemize}

\section{Related Work}
\label{sec:related}

In order to understand how to design the method for training a model to rewrite prompts, we look at previous work on LLM fine-tuning, prompt engineering, and existing prompt rewriting works. In addition, we look at the LLM-as-a-judge paradigm to explore alternative ways of evaluating and rewarding models.

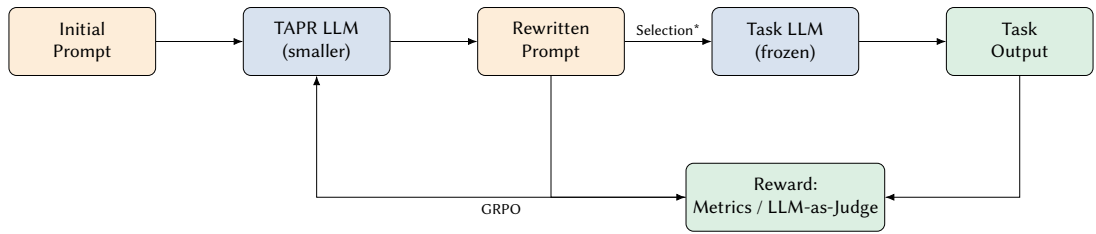
\begin{figure*}[h!]
    \centering
    %
    %
    \begingroup
      \definecolor{promptcol}{HTML}{F4A442}   
      \definecolor{modelcol}{HTML}{3C6EAF}    
      \definecolor{outputcol}{HTML}{4EAF6E}   
      \resizebox{0.9\textwidth}{!}{%
      \begin{tikzpicture}[
        font=\sffamily,
        >=Latex,
        node distance=1.6cm,
        column sep=2cm,
        box/.style     ={draw, rounded corners, align=center,
                         minimum width=2.7cm, minimum height=1.25cm},
        prompt/.style  ={box, fill=promptcol!20},
        model/.style   ={box, fill=modelcol!20},
        outnode/.style ={box, fill=outputcol!20}   
      ]
        \node[prompt]  (initial)                {Initial\\Prompt};
        \node[model,   right=of initial] (rewriter)  {TAPR LLM\\(smaller)};
        \node[prompt,  right=of rewriter] (rewritten) {Rewritten\\Prompt};
        \node[model,   right=of rewritten](taskLLM)  {Task LLM\\(frozen)};
        \node[outnode, right=of taskLLM]  (output)   {Task\\Output};
        \node[outnode, below=of taskLLM]  (reward)   {Reward:\\Metrics / LLM-as-Judge};

        \draw[->] (initial)   -- (rewriter);
        \draw[->] (rewriter)  -- (rewritten);
        \draw[->] (rewritten) -- node[above,pos=0.5]{\scriptsize Selection*}(taskLLM);
        \draw[->] (taskLLM)   -- (output);
        \draw[->] (rewritten) |- (reward);
        \draw[->] (output)    |- (reward);
        \draw[->] (reward)    -| node[below,pos=0.25]{\scriptsize GRPO} (rewriter);
      \end{tikzpicture}}%
    \endgroup

    \caption{Overview of the training pipeline for TAPR.
             \textbf{Orange} nodes are \emph{prompts} (inputs to models),
             \textbf{blue} nodes represent \emph{models} (trainable or frozen),
             and \textbf{green} nodes are \emph{outputs}—either the task answer
             or the scalar reward. Solid arrows show the forward data flow;
             the green reward is fed back via GRPO (curved arrow).}
    \label{fig:rl_pipeline}
\end{figure*}

\subsection{LLM Fine-tuning}
\label{sec:ft}

Supervised fine-tuning (SFT) is a general technique for adapting any pretrained LM to a labeled dataset; in the context of generative models it’s often called instruction-tuning, since the labels take the form of \textit{(prompt, response)} pairs \cite{wei2021finetuned}. Although SFT provides strong initialization for instruction following, it does not explicitly optimize for long-term objectives, such as user satisfaction or safety. Reinforcement learning from human feedback (RLHF) addresses this by training a scalar reward model on human preference judgments between model outputs, and then using Proximal Policy Optimization (PPO) to maximize that learned reward \cite{rlhf, ppo}. In the InstructGPT pipeline, a policy initialized by SFT is refined via RLHF, yielding substantial gains in helpfulness and coherence over purely supervised approaches \cite{instructgpt}.

More recently, Group Relative Policy Optimization (GRPO) was introduced which generalizes PPO by updating the policy based on a group of sampled trajectories, thereby removing explicit value function learning and enabling multi-task or multi-agent training regimes, leading to impressive performance for the DeepSeek R1 model \cite{grpo,deepseek}. Both GRPO and PPO have also been used to enhance other abilities, for example, Search-R1 \cite{jin2025searchr1trainingllmsreason} showed the effectiveness of both algorithms to make LLMs better in the use of search engines. Our work is the first to explore GRPO for enhancing prompt rewriting abilities.

\subsection{Prompt Engineering}
\label{sec:prompt}
Task framing has been proven to have a major impact on LLM performance. One of the first important breakthroughs for this was few-shot prompting, where providing just a handful of examples enables generalization without task-specific training \cite{brown2020language, boonstra2025prompt}. Chain-of-thought prompting (i.e., “Let’s think step by step”) elicits intermediate reasoning and boosts accuracy on arithmetic and logical tasks \cite{kojima2022zeroshot, boonstra2025prompt}. Few-shot chain-of-thought prompting remains essential for challenging reasoning benchmarks like GSM8K and StrategyQA, teaching models to break problems into steps \cite{wei2022chain}.

Clear, structured instructions also matter. Specifying output formats, especially JSON, improves consistency and makes results easier to parse \cite{shen2017style, hu2017toward, long2024llmsbiasedoutputformats}. Yet prompt design is still difficult, with no universally optimal formulation. Our approach tackles this by automatically rewriting prompts using prompt engineering principles.

\subsection{Prompt Rewriting}

Automated prompt engineering or prompt rewriting employs diverse optimization techniques to refine prompts that steer LLM behavior. AutoPrompt uses gradient‐based search over discrete token spaces to assemble sequences that elicit desired outputs, though often producing non‐interpretable prompts \cite{autop}. Evolutionary methods such as PromptBreeder treat prompt design as a population‐based search, applying crossover and mutation to candidate prompts and selecting offspring by performance \cite{breed}, but rely on carefully chosen initial prompts, limiting the search space.

Reinforcement learning offers a more sample‐efficient alternative by framing prompt generation as a sequential decision process. RLPrompt trains an agent to edit prompts token by token with rewards from task performance (e.g., classification accuracy, BLEU score), discovering prompts that generalize across datasets but remain hard to interpret \cite{rlprompt}. TEMPERA improves stability by defining a custom action space over high‐level edits (e.g., paraphrasing, task description insertion) and using Q‐learning to maximize a composite reward of performance and linguistic diversity \cite{tempera}, though its constrained actions may miss novel formulations.

PRewrite adopts PPO to fine‐tune LLMs (e.g., PaLM 2‐S/L \cite{palm}) to rewrite user prompts in a task‐aware manner \cite{prewrite}. Unlike token‐level policies or fixed edit spaces, it processes the entire prompt and outputs a rewritten version optimized for metrics such as exact match rate on the Natural Questions dataset or solution rate for GSM8K. Empirical results show that PRewrite improves task performance and yields more interpretable and concise prompts than RLPrompt, without TEMPERA’s constraints. However, it only uses automated metrics such as exact match accuracy, which can lead to prompts that do not optimize for real-world utility. Our work leverages LLM-based judging to account for that limitation.

\subsection{LLM-as-a-Judge}
\label{sec:llm-judge}
Traditional automatic evaluation metrics such as BLEU \cite{papineni2002bleu}, ROUGE \cite{lin2004rouge}, METEOR \cite{banerjee2005meteor}, and BERT-Score \cite{zhang2020bertscore} rely on n-gram overlap or embedding similarity with static reference texts, which can poorly reflect real‐world utility in open‐ended tasks such as summarization, dialogue, or long‐form question answering \cite{automatic}. These metrics often penalize valid paraphrases or novel content, and cannot account for pragmatics, coherence, or factuality beyond surface overlap. To address these shortcomings, the LLM-as-a-judge paradigm repurposes LLMs to score or rank model outputs by simulating human evaluators \cite{turn0academia8, gu2024survey}. By prompting the judge model with evaluation criteria such as relevance, fluency, and informativeness, it can produce scalar scores or pairwise preferences that align more closely with human judgments and adapt dynamically to new tasks without costly reference annotation.

Beyond zero-shot scoring, LLM judges have been integrated directly into model training loops. For example, RLAIF demonstrates that synthetic preferences generated by an auxiliary LLM can approximate human feedback for RLHF, enabling large‐scale reinforcement learning at a lower cost with minimal quality drop-off \cite{rlaif, bai2022constit}. This does not mean that using LLM judges leads to a perfect and faultless metric for every situation, but using LLMs to judge can often offer a better way to evaluate or calculate rewards for text generation, potentially leading to better LLM fine-tuning performance for various tasks. Our method is the first to integrate LLM-as-a-judge scores for rewards and evaluations to guide and improve automated prompt rewriting.

\begin{table*}[ht]
  \centering\small
  \renewcommand{\arraystretch}{1.15}
  \begin{tabularx}{\textwidth}{@{}l X X@{}}
    \toprule
    \textbf{Dataset} 
      & \textbf{Example} 
      & \textbf{Initial Prompt} \\
    \midrule
    Natural Questions (NQ) \citep{kwiatkowski-etal-2019-natural}
      & Who was the first president of the United States?
      & ``Answer the question.'' \citep{prewrite} \\[0.8ex]
    \hline
    HotpotQA \citep{yang-etal-2018-hotpotqa}
      & Which novel did Mary Shelley write that was published in 1818?
      & ``Answer the question based on the context.'' \\[0.8ex]
    \hline
    CNN/Daily Mail \citep{hermann2015teaching}
      & \textit{Article excerpt:} An Iraqi boy, badly burned, is flown to the U.S. for treatment funded by a charity.
      & ``Summarize the text.'' \\[0.8ex]
    \hline
    SciTLDR \citep{cachola-etal-2020-tldr}
      & \textit{Abstract excerpt:} We propose a 2-simplicial Transformer for deep RL and show it improves reasoning.
      & ``Summarize the text.'' \\[0.8ex]
    \hline
    GSM8K \citep{cobbe2021training}
      & Katy mixes sugar and water in a 7:13 ratio. If she used 120 units total, how many units of sugar were used?
      & ``SOLUTION.'' \citep{prewrite} \\
    \bottomrule
  \end{tabularx}
  \vspace{1ex}
  \caption{All the datasets used in the experiments, example samples from them, and initial baseline prompt for rewriting we are using for them. }
  \label{tab:datasets}
\end{table*}
\section{Task-Aware Prompt Rewriter}
\label{sec:methods}

In this section, we describe the details of our method for creating the Task-Aware Prompt Rewriter (TAPR). The overall training method for TAPR can be seen in Figure \ref{fig:rl_pipeline}. The training procedure works as follows: for each sample, we rewrite the initial prompt using the TAPR LLM, then we give the rewritten prompt to the Task LLM, which performs the task, and finally, we calculate the rewards using both the rewritten prompt and the task. Based on the reward values, the model is trained with a reinforcement learning algorithm.

\subsection{Problem Formulation}

TAPR is a smaller LLM that reformulates queries for a bigger frozen Task LLM. We formulate the problem similarly to \cite{prewrite} and \cite{Li2025Survey}. Let $\mathcal{P}$ denote the space of textual prompts. We consider the TAPR LLM $\pi_\theta:\mathcal{P}\to\mathcal{P}$ that given an original prompt $p\in\mathcal{P}$ outputs a revised prompt $\tilde p=\pi_\theta(p)$. The revised prompt is then fed into a frozen Task LLM $L$, yielding an output $y=L(\tilde p)$. Let $y^*$ be the desired (ground-truth) output corresponding to input $p$. We define two reward functions: a task performance reward $T(y,y^*)$ (such as LLM-as-a-judge score or accuracy), measuring how well $y$ matches $y^*$, and a prompt quality reward $Q(\tilde p)$ that encourages the rewritten prompt to be accurate and follow common prompt engineering principles.

For reinforcement learning, we treat $\pi_\theta$ as a policy over the discrete prompt space. We define a combined reward 
\[
R(y,\tilde p) = \alpha\,T(y,y^*) + \beta\,Q(\tilde p),
\]
where $\alpha,\beta>0$ are weighting coefficients. The reinforcement learning objective is to maximize the expected reward under the data distribution $\mathcal{D}$ of prompts:
\[
\mathcal{J}_{\mathrm{RL}}(\theta)
= \mathbb{E}_{\substack{
    p \sim \mathcal{D}\\
    \tilde p \sim \pi_\theta(\,\cdot\mid p)
}}
\bigl[R\bigl(L(\tilde p),\,\tilde p\bigr)\bigr].
\]
We optimize the expectation via policy gradient methods such as PPO \cite{ppo} or GRPO \cite{grpo}, treating prompts as policies. This formulation follows the optimization perspective of prompt tuning (treating prompts as differentiable policies and maximizing downstream task metrics).

\subsection{Rewards}
\label{sec:rewards}

We train TAPR  using the GRPO algorithm \cite{grpo}, leveraging its recent success in advancing reasoning and other abilities within language models. It has also been used more commonly with verifiable rewards such as accuracy rather than rewards coming from a model trained on preference data, such as PPO and DPO \cite{dpo, ppo, grpo}.

The TAPR model gets the initial prompts as input, and rewrites them. Then both the prompts and responses from the Task LLM are evaluated. The initial prompt can be a general instruction such as \textit{"Answer the question"}, or it can also include the question and context if one exists. Example meta-prompt is given in Appendix \ref{app:inst-only-meta}. We use two types of rewards with equal weights: Task LLM performance rewards and the prompt quality reward. For the former, we use scores from an LLM-as-a-judge evaluator rather than traditional metrics such as exact match accuracy and ROUGE due to their limitations. 

This approach enables the evaluation of semantic correctness, accounting for cases in which LLMs generate responses that are more verbose than the ground-truth labels but nonetheless accurate\footnote{We verified the effectiveness of this approach through running an internal test, and we evaluated that only 1 out of 100 of the judgments on the Natural Questions were incorrect.}.

The \textbf{prompt quality reward }is a score on a scale of 0 to 5 coming from an LLM judge based on whether the rewritten prompt matches the meaning of the initial instruction and improves it according to common prompt engineering principles. This decreases the chances of generating prompts with drastic inaccuracies (for example, by completely changing the meaning of the initial instruction or even just essentially repeating the rewriting instruction). As LLM-as-a-judge usage can lead to unwanted bias, we try to mitigate this by using two different models as judges.

\subsection{Prompt Selection Mechanism}
To further improve rewriting performance, we introduce a \textbf{Selection} mechanism, where the trained TAPR model selects the most promising prompt from multiple generated options sampled with a higher temperature during inference. We hypothesize that, while this selection is not the main task of the TAPR LLM, the trained model has gathered enough knowledge about prompt engineering to evaluate the quality of prompts in addition to generating them. We compare this approach to an approach where prompts are generated by models using a temperature set to zero.
\section{Experimental Setup}
\label{sec:exp_setup}

We evaluate the performance of TAPR across different tasks, including question answering, summarization, and arithmetic reasoning (GSM8K) \cite{cobbe2021training}. Although we also report standard metrics for these datasets, our primary focus is on LLM-as-a-judge evaluations for question answering and summarization, as these better address the limitations of conventional evaluation methods as detailed in Section \ref{sec:llm-judge}.

In Table \ref{tab:datasets}, we show examples from each dataset we are using alongside the initial baseline prompts we are using for rewriting.

We evaluate prompt rewriting performance across a variety of models, tasks, and training methods. Each experiment involves two distinct LLMs: the TAPR LLM, which generates improved prompts, and the Task LLM, which solves the downstream task using these rewritten prompts. Our main experiments use LLaMA-3.1-8B-Instruct \cite{touvron2023llama} as the Task LLM and compare two open-source TAPR LLMs: Phi-4-mini-instruct \cite{microsoft2024phi4}, LLaMA-3.2-3B-Instruct \cite{meta2024llama}. Results with Qwen3-4B \cite{qwen} are in Appendix \ref{sec:nq_full}.

\noindent\textbf{Reinforcement Learning Training Setup and Stopping Criterion:}
We train each TAPR LLM using reinforcement learning, precisely the GRPO~\cite{grpo} algorithm, until convergence, defined by a lack of improvement in the 25-step moving average of the reward for 100 consecutive steps. This stopping criterion prevents excessive training once the model’s performance plateaus. The rewards used are specified in Section \ref{sec:rewards}. To reduce bias and stabilize the reward signal, we average the LLM-as-a-judge scores from two different judge models (LLaMA-3.1-8B-Instruct and GPT-4o-mini). 

Hyperparameters for training are listed in Appendix \ref{sec:hyper}. To maximize experimental coverage rather than repeat identical runs with different seeds, each training and evaluation is conducted only once.

We observed that the TAPR LLM might generate responses containing information other than the rewritten queries, such as explanations for the rewriting decisions made or responses to the initial task. This might lead to the Task LLM being unable to answer the query. Therefore, we instruct TAPR to use a JSON format. Apart from this change, we closely follow most templates and initial prompts used in PRewrite \cite{prewrite}. This gives us reliable baselines and a starting point to improve any other prompts. A full example of a meta-prompt can be seen in Appendix \ref{app:inst-only-meta}.

\noindent\textbf{Evaluation Protocol:} Final evaluations are conducted on 1,000 samples from the validation or test split of each dataset. During evaluation, we use GPT-4o-mini as the sole judge model to score answers (avoiding any bias from using the same model for generation and judgment). In our internal tests (Section~\ref{sec:rewards}), verdicts from GPT-4o-mini were nearly perfectly accurate on the question-answering tasks.

\noindent\textbf{Decoding Strategies:} For consistency and reproducibility, we use a temperature of zero for the Task LLM’s answer generation and for all judge evaluations. For TAPR’s generation, we compare two strategies: (a) \emph{TAPR}, where the rewriter uses temperature zero and returns a single fixed rewrite; and (b) \emph{TAPR + Selection}, where the rewriter samples five candidate prompts using a moderate temperature (0.5) and then selects the best candidate based on its judgment. We chose five samples to balance diversity and computational efficiency.

\noindent\textbf{Ablation Studies:} To isolate the contribution of the GRPO algorithm, prompt quality reward, and using SFT before TAPR training, we also conduct ablation studies using the Phi-4-mini-instruct as TAPR LLM with LLaMA-3.1-8B-Instruct as the Task LLM. We consider Phi-4-mini-instruct to be a capable model for its size, which responds well to reinforcement learning training in our experiments. LLaMA-3.1-8B-Instruct was chosen as the Task LLM as it is a slightly bigger LLM than Phi-4-mini-instruct, while not being from the same model family.

We also perform experiments using full-prompt rewriting, and alternative Task LLMs, and report these results in the Appendices \ref{sec:fpr} and \ref{sec:alter}. To assess how task-aware and generalizable our method is, we also report experimental results across tasks in Appendix \ref{sec:gen}.

\section{Experimental Results}

We compare four conditions for each base model in each experiment:
\begin{itemize}
  \item \textbf{Baseline Prompt}: The unmodified initial instruction.
  \item \textbf{Base Model}: Rewriting with the base version of the LLM before training.
  \item \textbf{TAPR}: Rewriting after training the rewriter LLM with our method.
  \item \textbf{TAPR + Selection}: Generating multiple rewritten prompts after training and selecting one of them.
\end{itemize}

This allows us to answer our research question and analyze whether our method is effective in enhancing the prompt rewriting abilities of LLMs. All results are reported based on the quality of the Task LLM responses.

\subsection{Impact of Prompt Rewriting on Different Tasks}
\label{sec:results}
Here, we evaluate whether TAPR enhances the performance of LLMs in downstream tasks.

\noindent\paragraph{\textbf{Question Answering}} Table \ref{tab:nq} reports accuracy based on LLM judgments on the Natural Questions (NQ) dataset. In Appendix Table \ref{tab:nq_full}, we demonstrate that standard exact-match metrics often yield low scores not aligned with the actual answer quality, suggesting that LLM-as-a-judge evaluation can capture practical performance better.
For Phi-4-mini-instruct and LLaMA-3.2-3B-Instruct, both deterministic rewrites and TAPR + Selection variants outperform their Base Model versions and surpass the Baseline prompt's accuracy.

\begin{table}[h!]
  \centering{!}{%
    \begin{tabular}{@{} l l c @{}}
      \toprule
      \textbf{TAPR LLM} & \textbf{Rewriting Variant} & \textbf{Accuracy $\uparrow$} \\
      \midrule
      Baseline prompt & – & 56.30 \\
      \cdashline{1-3}[.5pt/2pt]
      Phi-4-mini-instruct & Base Model & 48.80 \\
      Phi-4-mini-instruct & TAPR & \textbf{59.20} \\
      Phi-4-mini-instruct & TAPR + Selection & 57.30 \\
      \cdashline{1-3}[.5pt/2pt]
      LLaMA-3.2-3B-Instruct & Base Model & 58.30 \\
      LLaMA-3.2-3B-Instruct & TAPR & \textbf{62.20} \\
      LLaMA-3.2-3B-Instruct & TAPR + Selection & 59.10 \\
      \bottomrule
    \end{tabular}%
  }
  \caption{NQ dataset results, each value is an accuracy percentage (\%), measured by GPT-4o-mini, reflecting answer correctness for each prompt variant.}
  \label{tab:nq}
\end{table}

\begin{table}[h!]
  \centering{%
    \begin{tabular}{@{} l l c @{}}
      \toprule
      \textbf{TAPR LLM} & \textbf{Rewriting Variant} & \textbf{Accuracy $\uparrow$} \\
      \midrule
      Baseline prompt & – & 53.20 \\
      \cdashline{1-3}[.5pt/2pt]
      Phi-4-mini-instruct & Base Model & 53.20 \\
      Phi-4-mini-instruct & TAPR & \textbf{56.70} \\
      Phi-4-mini-instruct & TAPR + Selection & 53.16 \\
      \cdashline{1-3}[.5pt/2pt]
      LLaMA-3.2-3B-Instruct & Base Model & 53.15 \\
      LLaMA-3.2-3B-Instruct & TAPR & \textbf{56.00} \\
      LLaMA-3.2-3B-Instruct & TAPR + Selection & 55.00 \\
      \bottomrule
    \end{tabular}%
  }
  \caption{HotpotQA dataset results, each value is an accuracy percentage (\%), measured by GPT-4o-mini, reflecting answer correctness for each prompt variant.}
  \label{tab:hotpotqa}
\end{table}

Example prompts illustrating the rewriting for Natural Questions are provided in Table~\ref{tab:nq_prompts}. The results show that the Phi-4-mini-instruct’s base model rewrite is focused on describing the process, while RL fine-tuning yields more targeted and concise instructions. The best-performing prompt, generated by Llama 3.2-3B-Instruct TAPR, combines accuracy requirements with explicit formatting and clarity constraints.

\begin{table*}[h!]
  \centering
  \makebox[\textwidth][c]{%
    \begin{tabular}{@{} l l p{8.2cm} @{}}
      \toprule
      \textbf{TAPR LLM} & \textbf{Rewriting Variant} & \textbf{Prompt} \\
      \midrule
      -- & Initial Prompt & Answer the question \\
      \hline
      Phi-4-mini-instruct & Base Model & Provide a detailed explanation of the process you used to answer the question, including any relevant examples or illustrations. \\
      \hline
      Phi-4-mini-instruct & TAPR & Answer the question and provide a brief explanation for your answer. \\
      \hline
      LLaMA-3.2-3B-Instruct & TAPR & Provide a well-supported answer to the question in a concise paragraph (approximately 50--75 words) or a list of 3--5 bullet points. Ensure your response is accurate and relevant, and include evidence or examples to support your answer. Use clear, concise language and avoid ambiguity. \\
      \bottomrule
    \end{tabular}
  }
  \caption{Example prompts for the Natural Questions dataset. Rows compare the baseline, the pre‑RL rewrites, and the TAPR outputs.}
  \label{tab:nq_prompts}
\end{table*}

\noindent\paragraph{\textbf{Summarization}} For summarization, we report LLM-as-a-judge scores out of 5 in Tables~\ref{tab:cnndm} and \ref{tab:scitldr} on the CNN/Daily Mail and SciTLDR datasets.

The baseline summarization instruction performs noticeably worse than several rewritten prompt variants. On both datasets, Phi-4-mini-instruct trained variants yield the highest scores.

The best overall scores come from TAPR and TAPR+Selection variants of Phi-4-mini-instruct on the datasets, respectively. These prompts are shown in Table~\ref{tab:summary_prompts}. Both prompts are well-matched to their respective dataset styles and provide strong length constraints, leading to more focused summaries. In contrast, over-specified prompts (such as LLaMA-3.2-3B-Instruct’s multi-sentence summary format) can degrade performance by instructing the use of formats that do not match the reference summaries. LLaMA results overall are weaker here, which could be due to the hyperparameters being optimized based on experiments with Phi-4-mini.

\begin{table}[h!]
  \centering{%
    \begin{tabular}{@{} l l r @{} }
      \toprule
      \textbf{TAPR LLM} & \textbf{Rewriting Variant} & \textbf{Judge Score (1-5) $\uparrow$} \\
      \midrule
      Baseline prompt & – & 3.373 \\
      \cdashline{1-3}[.5pt/2pt]
      Phi-4-mini-instruct & Base Model & 3.772 \\
      Phi-4-mini-instruct & TAPR & \textbf{3.782} \\
      Phi-4-mini-instruct & TAPR + Selection & 3.774 \\
      \cdashline{1-3}[.5pt/2pt]
      LLaMA-3.2-3B-Instruct & Base Model & \textbf{3.794} \\
      LLaMA-3.2-3B-Instruct & TAPR & 3.435 \\
      LLaMA-3.2-3B-Instruct & TAPR + Selection & 3.411 \\
      \bottomrule
    \end{tabular}%
  }
  \caption{CNN/DM dataset summarization results, each value is an LLM-as-a-judge score out of 5, as rated by GPT-4o-mini, reflecting summary quality for each prompt variant.}
  \label{tab:cnndm}
\end{table}

\begin{table}[h!]
  \centering{%
    \begin{tabular}{@{} l l r @{} }
      \toprule
      \textbf{TAPR LLM} & \textbf{Rewriting Variant} & \textbf{Judge Score (1-5) $\uparrow$} \\
      \midrule
      Baseline prompt & – & 3.182 \\
      \cdashline{1-3}[.5pt/2pt]
      Phi-4-mini-instruct & Base Model & 3.802 \\
      Phi-4-mini-instruct & TAPR & 3.710 \\
      Phi-4-mini-instruct & TAPR + Selection & \textbf{3.869} \\
      \cdashline{1-3}[.5pt/2pt]
      LLaMA-3.2-3B-Instruct & Base Model & \textbf{3.749} \\
      LLaMA-3.2-3B-Instruct & TAPR & 3.107 \\
      LLaMA-3.2-3B-Instruct & TAPR + Selection & 3.041 \\
      \bottomrule
    \end{tabular}%
  }
  \caption{SciTLDR dataset summarization results, each value is an LLM-as-a-judge score out of 5, as rated by GPT-4o-mini, reflecting summary quality for each prompt variant.}
  \label{tab:scitldr}
\end{table}

\noindent\paragraph{\textbf{Pairwise Win-Rate Results for Summarization}} To complement the numeric scores, we also report win-rates comparing summaries generated from the baseline prompts with the ones generated with the rewritten prompts in Table~\ref{tab:winrate_combined}. These results confirm that TAPR variants are often preferred over both the baseline and base model outputs. None of the TAPR models had a win-rate lower than 40\%; notably, Phi-4-mini-instruct achieves almost 85\% win-rates on SciTLDR, while LLaMA-3.2-3B achieves between 47\%--63\% win-rates despite the numeric score showing worse performance. We do observe a strong bias of the LLM-as-a-judge towards the second shown option, which is why we shuffled outputs randomly for each sample at comparison time.

\begin{table*}[h!]
  \centering
  \normalsize
  \setlength{\tabcolsep}{8pt}       
  \renewcommand{\arraystretch}{1.2} 

  \begin{tabular}{@{} l l c c c c @{}}
    \toprule
    \textbf{Dataset} & \textbf{TAPR LLM}
      & \multicolumn{2}{c}{\textbf{vs. Baseline}}
      & \multicolumn{2}{c}{\textbf{vs. Base}} \\
    \cmidrule(lr){3-4}\cmidrule(lr){5-6}
      & & \textbf{Overall $\uparrow$} & \textbf{Placed 1st/2nd $\uparrow$}
        & \textbf{Overall $\uparrow$} & \textbf{Placed 1st/2nd $\uparrow$} \\
    \midrule
    \multirow{2}{*}{CNN/Daily Mail}
      & Phi-4-mini   & 46.11 & 18.13/72.99 & 40.66 & 20.41/59.69 \\
      & LLaMA-3.2-3B & 62.89 & 37.45/89.09 & 67.03 & 39.07/94.28 \\
    \midrule
    \multirow{2}{*}{SciTLDR}
      & Phi-4-mini   & 84.58 & 75.23/94.88 & 84.42 & 86.09/82.80 \\
      & LLaMA-3.2-3B & 47.97 & 37.11/59.53 & 50.73 & 39.88/62.89 \\
    \bottomrule
  \end{tabular}

  \caption{Win–rate comparison on CNN/Daily Mail and SciTLDR datasets. “Overall” is the percentage of cases where TAPR was preferred by GPT-4o-mini; “Placed 1st/2nd” reports preferences when shown first vs.\ second, compared to both baseline and base model prompts.}
  \label{tab:winrate_combined}
\end{table*}

\begin{table*}[h!]
\centering
\makebox[\textwidth][c]{%
\begin{tabular}{@{} l l l p{8.2cm} @{}}
\toprule
\textbf{Dataset} & \textbf{TAPR LLM} & \textbf{Rewriting Variant} & \textbf{Prompt} \\
\midrule
CNN/DM, SciTLDR & -- & Initial Prompt & Summarize the text \\
\hline
CNN/DM & Phi-4-mini-instruct & TAPR & Please provide a concise summary of the provided text, aiming for a succinct overview suitable for a general audience. The summary should be no longer than one-third of the original text length, maintaining a neutral and informative tone. Ensure that the key points and main ideas are preserved while eliminating any redundant or less critical information. \\
\hline
SciTLDR & LLaMA-3.2-3B-Instruct & TAPR & Condense the given text into a 3- to 5-sentence summary, highlighting key points and main ideas. Present the summary in a neutral, third-person voice, without using the author's name or any personal pronouns. For example, A recent study found that [key finding]... \\
\bottomrule
\end{tabular}
}
\caption{Example prompts for the CNN/DM and SciTLDR datasets.
  Rows compare the baseline, the pre‑RL rewrites, and the TAPR outputs.}
\label{tab:summary_prompts}
\end{table*}

\noindent\paragraph{\textbf{Arithmetic Reasoning (GSM8K)}}

For the arithmetic reasoning task with the GSM8K dataset, we report accuracy based on the last integer found in the model’s response, matching the reward used for training. While this approach may occasionally misattribute the answer if multiple numbers are present, it also serves as a strong indicator of whether prompts guide models to output formats suited to the evaluation method. Results can be seen in Table \ref{tab:gsm8k}.

The baseline ``SOLUTION'' prompt is surprisingly robust and yields high accuracy relatively to many of the more detailed rewritten prompts. Nevertheless, the Phi-4-mini-instruct TAPR variant surpasses it, while also exhibiting more than a 70\% improvement over the base model. 

Table~\ref{tab:gsm8k_prompts} includes example prompt rewrites for GSM8K. The Phi-4-mini-instruct’s base model rewrite asks for a Python function to sum integers, likely leading to confusion for the Task LLM and inability to produce the final correct answer with the limited maximum response tokens. This leads to poor performance, which all other rewrites clearly surpass.

In contrast, the TAPR variant produces a highly relevant, task-specific chain-of-thought prompt that encourages explicit step-by-step reasoning, justifications for each step, and a clear final answer. This style directly supports the kind of solutions needed for GSM8K, resulting in substantial performance improvements.

Despite improvements, neither the summarization nor the arithmetic reasoning results show substantial improvement over the base model in the former or the short general prompt in the latter task, with some trainings and variants leading to worse task performance. In addition to the potential issue with hyperparameters, this highlights how RL training for a task that is not measured directly based on the output of the model being trained is challenging, and not always reliable. Furthermore, despite significant research, prompt engineering is still an open problem where we can not exactly know which prompt leads to the optimal performance for the Task LLM. Using Selection leads to improvement sometimes, but overall, we find no strong evidence that training for prompt rewriting consistently enhances a model’s ability to judge and select among candidate prompts.

\begin{table}[h!]
  \centering{%
    \begin{tabular}{@{} l l c @{}}
      \toprule
      \textbf{TAPR LLM} & \textbf{Rewriting Variant} & \textbf{Accuracy $\uparrow$} \\
      \midrule
      Baseline – ``SOLUTION'' & -- & 82.38 \\
      \cdashline{1-3}[.5pt/2pt]
      Phi-4-mini-instruct & Base Model & 11.91 \\
      Phi-4-mini-instruct & TAPR & \textbf{83.60} \\
      Phi-4-mini-instruct & TAPR + Selection & \textbf{83.60} \\
      \cdashline{1-3}[.5pt/2pt]
      LLaMA-3.2-3B-Instruct & Base Model & 74.80 \\
      LLaMA-3.2-3B-Instruct & TAPR & \textbf{82.80} \\
      LLaMA-3.2-3B-Instruct & TAPR + Selection & 81.80 \\
      \bottomrule
    \end{tabular}%
  }
  \caption{GSM8K dataset results comparing the Baseline against the base models and TAPR variants. Each value is an accuracy percentage (\%), reflecting answer correctness for each prompt variant.}
  \label{tab:gsm8k}
\end{table}

\begin{table*}[h!]
  \centering
  \makebox[\textwidth][c]{%
    \begin{tabular}{@{} l l p{8.2cm} @{}}
      \toprule
      \textbf{TAPR LLM} & \textbf{Rewriting Variant} & \textbf{Prompt} \\
      \midrule
      -- & Initial Prompt & SOLUTION \\
      \hline
      Phi-4-mini-instruct & Base Model & Please provide a detailed solution to the following problem: Given a list of integers, write a function in Python that returns the sum of all even numbers in the list... \\
      \hline
      Phi-4-mini-instruct & TAPR & Provide a step-by-step solution to the problem, including intermediate steps, justifications for each step, and a clear explanation of the final answer. \\
      \bottomrule
    \end{tabular}
  }
  \caption{Example prompts for the GSM8K dataset. Rows compare the baseline, the pre‑RL rewrites, and the TAPR outputs.}
  \label{tab:gsm8k_prompts}
\end{table*}

\subsection{GRPO vs PPO}
\label{sec:ppo}

Here we show results comparing using the GRPO algorithm for our method to the PPO algorithm used in PRewrite \cite{prewrite}. Table~\ref{tab:ppo} summarizes the results on both the GSM8K and Natural Questions datasets. We observe that GRPO reliably improves the training reward and downstream task accuracy for both datasets. Depending on the hyperparameters, PPO training either results in stagnant rewards or leads to model collapse, with degenerate outputs such as repetitive symbols or nonsense token.

Although prior work in PRewrite \cite{prewrite} showed that PPO can be effective for prompt rewriting training, differences in model architecture, hyperparameter choices, and using a critic head instead of a full critic model may contribute to the discrepancies observed here. Due to these challenges and the lack of access to the exact implementation in PRewrite \cite{prewrite}, we opted to use GRPO for all other experiments, and couldn't include a PRewrite baseline to compare to.

\begin{table}[!t]
  \centering
  \small
  \setlength{\tabcolsep}{6pt}
  \begin{tabularx}{\linewidth}{@{} l L c c @{}}
    \toprule
    \textbf{TAPR LLM} & \textbf{Rewriting Variant} & \textbf{NQ} $\uparrow$ & \textbf{GSM8K} $\uparrow$ \\
    \midrule
    Baseline & -- & 56.30 & 82.38 \\
    \cdashline{1-4}[.5pt/2pt]
    Phi-4-mini-instruct & Base Model & 48.80 & 11.91 \\
    Phi-4-mini-instruct & TAPR with GRPO & \textbf{59.20} & \textbf{83.60} \\
    Phi-4-mini-instruct & TAPR with PPO — Con. & 35.90 & 71.10 \\
    Phi-4-mini-instruct & TAPR with PPO — Mod. & 4.00 & 75.20 \\
    Phi-4-mini-instruct & TAPR with PPO — Agg. & 52.10 & 78.40 \\
    \bottomrule
  \end{tabularx}
  \caption{Comparison of performance using GRPO and three PPO variants for training the TAPR on NQ and GSM8K (Accuracy \%). "Con." (“Conservative”), "Mod" (“Moderate”), and "Agg." (“Aggressive”) are PPO hyperparameter settings differing mainly in learning rate and KL-divergence control values.}
  \label{tab:ppo}
\end{table}

\newpage

\subsection{SFT impact on TAPR}
In this section, we see how effective is using Supervised Fine-Tuning (SFT) prior to reinforcement learning. 

We conduct SFT for 5 epochs, with data coming from various sources showing rewrites of prompts, chain-of-thought prompts and responses, and ratings to prompts given by humans \cite{10k_prompts_ranked,prompt_optimization_dataset}. In addition, we instructed GPT-4o-mini \cite{openai2024gpt4omini} to generate prompts using common prompt engineering techniques for queries coming from the datasets used in the RL phase. We compare training on our internally generated dataset with training on publicly available datasets combined with our internally generated dataset. As seen in Table \ref{tab:sft_comparison}, SFT improves the base performance of Phi-4-mini-instruct, particularly when conducted with both internally generated and external data. However, this improvement does not translate into enhanced RL training outcomes.

\begin{table}[!t]
  \centering
  \small
  \setlength{\tabcolsep}{6pt}
  \begin{tabularx}{\linewidth}{@{} l L c c @{}}
    \toprule
    \textbf{TAPR LLM} & \textbf{Rewriting Variant} & \textbf{NQ} $\uparrow$ & \textbf{GSM8K} $\uparrow$ \\
    \midrule
    Baseline & – & 56.30 & 82.38 \\
    \cdashline{1-4}[.5pt/2pt]
    Phi-4-mini-instruct & Base Model & 48.80 & 11.91 \\
    Phi-4-mini-instruct & SFT with generated data & 52.50 & 35.34 \\
    Phi-4-mini-instruct & Full SFT & 54.00 & 71.00 \\
    Phi-4-mini-instruct & TAPR & \textbf{59.20} & \textbf{83.60} \\
    Phi-4-mini-instruct & TAPR after SFT & 55.50 & 70.50 \\
    \bottomrule
  \end{tabularx}
  \caption{Comparison of TAPR performance on the NQ and GSM8K datasets, evaluating the impact of supervised fine-tuning (SFT) with internally generated data, externally sourced data (“Full SFT”), and reinforcement learning (RL), both with and without prior SFT. Results are reported as accuracy percentages (\%).}
  \label{tab:sft_comparison}
\end{table}

\subsection{Impact of the Prompt Quality Reward}
\label{sec:pq}
As an additional ablation study, we evaluate the effect of using the prompt quality reward.
    
\begin{figure}[h]
  \centering
  \includegraphics[width=0.5\textwidth]{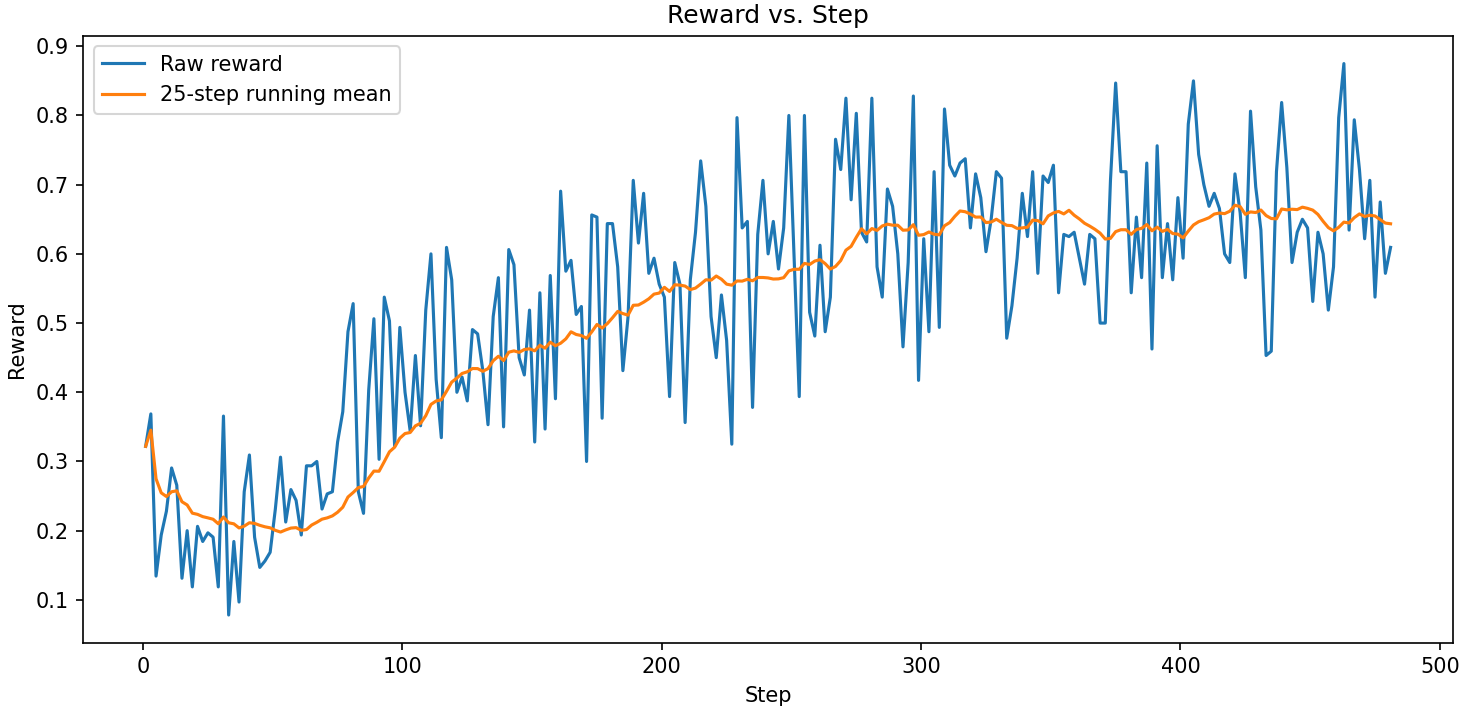}%
  \hfill
  \includegraphics[width=0.5\textwidth]{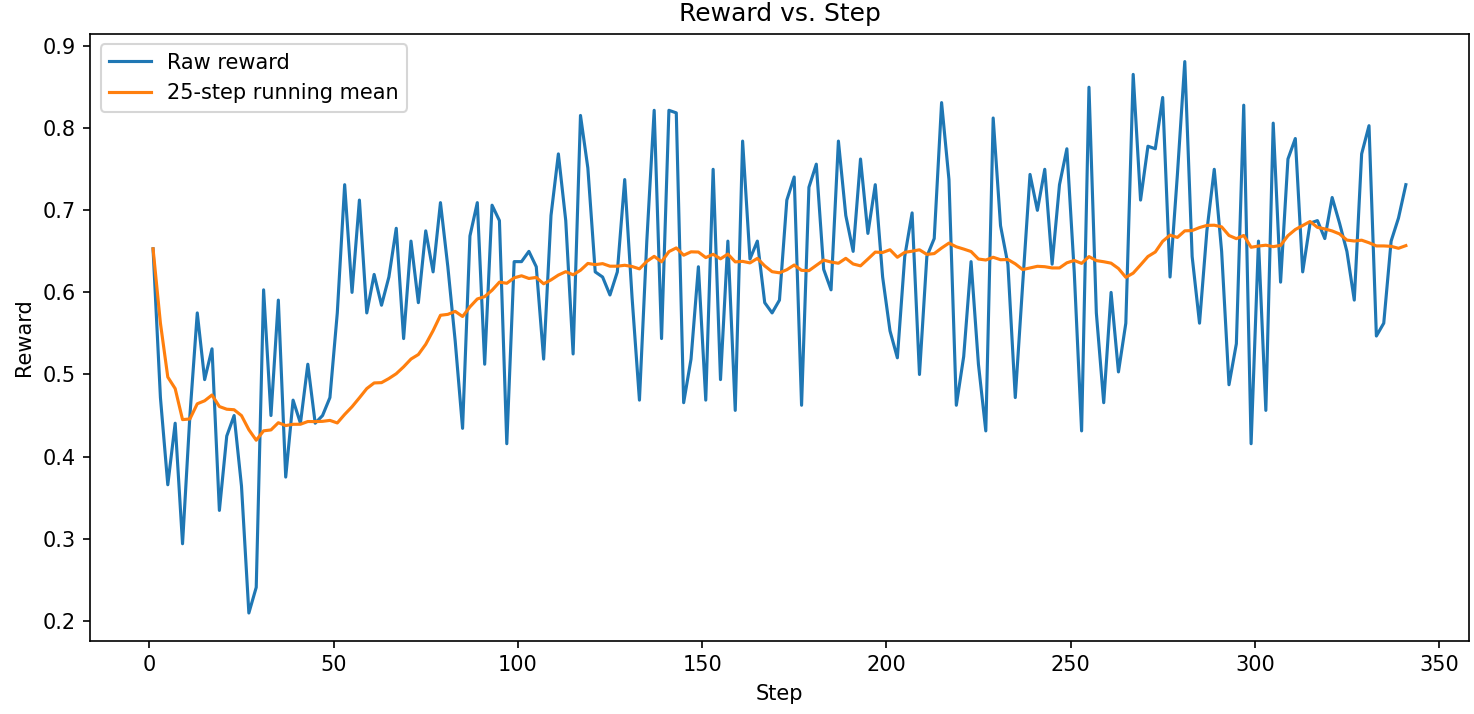}
  \caption{Training rewards comparison for the TAPR on the NQ dataset. The upper panel shows the raw per‐step reward (blue) and its 25‐step moving average or running mean (orange) with our method, and the bottom panel shows the training without the prompt quality reward.}
  \label{fig:sft_nq_training_comp}
\end{figure}

\begin{figure}[h]
  \centering
  \includegraphics[width=0.5\textwidth]{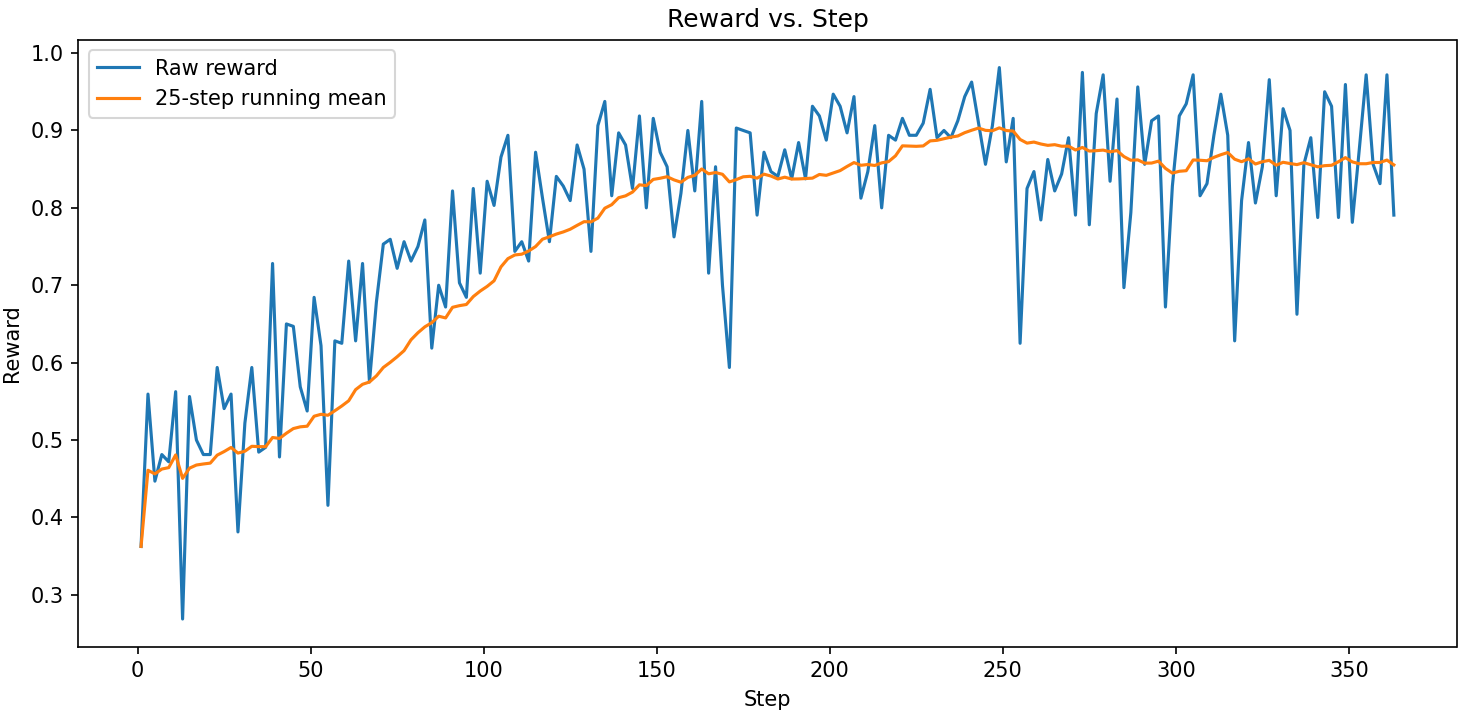}%
  \hfill
  \includegraphics[width=0.5\textwidth]{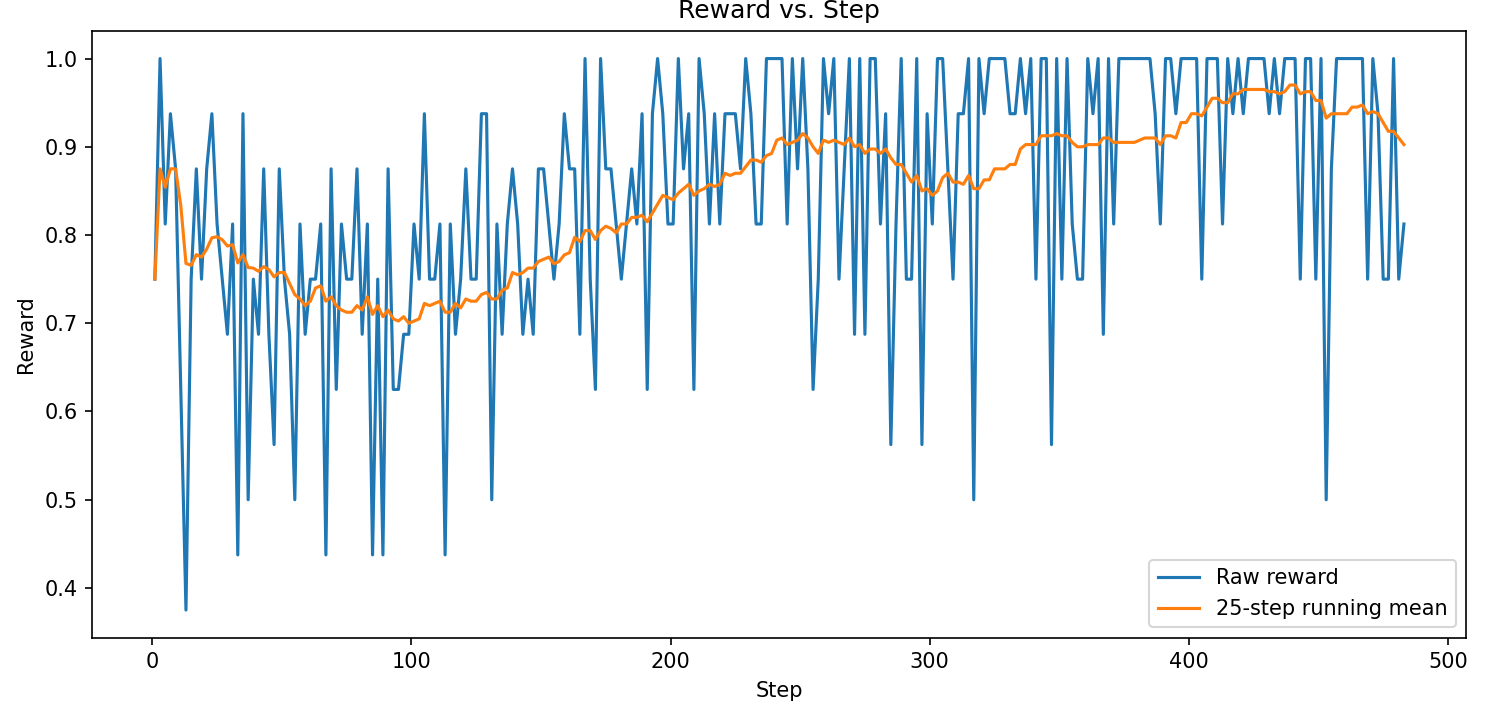}
  \caption{Training rewards comparison for the TAPR on the GSM8K dataset. The upper panel shows the raw per‐step reward (blue) and its 25‐step moving average or running mean (orange) with our method, and the bottom panel shows the training without the prompt quality reward.}
  \label{fig:sft_gsm_training_comp}
\end{figure}

As shown in Table \ref{tab:pq} and Figures \ref{fig:sft_nq_training_comp} and \ref{fig:sft_gsm_training_comp}, integrating the prompt quality reward improves both the accuracy of the downstream task and the convergence of training. In contrast, omitting this reward severely limits training progress. Without it, the model rewrites remain essentially unchanged from the initial prompts (``Answer the question." for Natural Questions and ``Provide the solution." for GSM8K), indicating negligible learning.

\begin{table}[h!]
  \centering
  \begin{tabular}{@{} l l c c @{} }
    \toprule
    \textbf{TAPR LLM} & \textbf{Rewriting Variant} & \textbf{NQ} $\uparrow$ & \textbf{GSM8K} $\uparrow$ \\
    \midrule
    Baseline & -- & 56.30 & 82.38 \\
    \cdashline{1-3}[.5pt/2pt]
    Phi-4-mini & Base Model & 48.80 & 11.91 \\
    Phi-4-mini & TAPR & \textbf{59.20} & \textbf{83.60} \\
    Phi-4-mini & TAPR without PQ Reward & 57.20 & 82.80 \\
    \bottomrule
  \end{tabular}
  \caption{Comparison of performance with and without prompt quality reward on the NQ and GSM8K datasets using Phi-4-mini-instruct as the base model (Accuracy \%).}
  \label{tab:pq}
\end{table}

\section{Discussion}
\label{sec:limitations}

\subsection{Performance Discussion}

Our results indicate that, although the TAPR method can produce improved prompt rewrites and boost performance, these gains are not always consistent. In several cases, even simple or generic prompts achieve competitive or better task results than more highly optimized rewrites. This underscores the difficulty of reinforcement learning for prompt engineering: learning a rewriting policy based on rewards that do not directly reference the Task LLM’s outputs is inherently challenging.

While the prompt quality reward correlates with better training rewards and often improved task scores, as seen in Section \ref{sec:pq}, much of the reward improvement may reflect the model’s ability to satisfy prompt quality criteria rather than the end-task metrics. Thus, increases in training reward do not always translate directly into downstream accuracy gains.

Prompt engineering is, by nature, a challenging problem. The relationship between prompt structure and LLM performance is complex and sometimes unpredictable. During RL training, models can receive mixed signals: for some samples, even a suboptimal or unrelated prompt may elicit a correct answer, while a carefully crafted prompt might not.

Further complicating matters, RL for text generation is inherently unstable. Since the reward is based on the entire output rather than single decisions, training is highly sensitive to hyperparameter choices. As a result, identifying robust configurations can require significant trial and error. In our experiments, this issue was apparent for LLaMA-3.2-3B-Instruct, likely due to the hyperparameters being optimized based on training Phi-4-mini-instruct. In summary, while the Task-Aware Prompt Rewriter approach shows promise, reliably optimizing prompts through RL remains an open challenge.

\subsection{Prompt Selection Mechanism}

The Selection mechanism sometimes improves over basic TAPR outputs, but these gains are inconsistent. As a result, we find no strong evidence that training for prompt rewriting consistently enhances a model’s ability to judge and select among candidate prompts. This is likely due to the fact that our training only improves performance with a specific prompt without intending to improve any other abilities. Given the additional computational cost, 0-temperature generation may be preferable for simplicity, despite the risk of a single failed prompt affecting all outputs. For instance, in one Phi-4-mini-instruct training run on CNN/Daily Mail, test-time selection modestly outperformed the base model, yet 0-temperature generation produced a prompt such as “Could you tell me a light-hearted, family-friendly joke, preferably related to technology or everyday life?”, a complete mismatch that caused the Task LLM to generate jokes rather than summaries for all inputs.

\subsection{LLM-as-a-Judge Evaluation}

Although LLM-as-a-judge evaluation is central to this work, we acknowledge its limitations and potential biases. For example, as observed with GPT-4o-mini in Section \ref{sec:results}, the model exhibited a strong preference for whichever response was presented second. Still, we argue that this approach offers a more realistic assessment of open-ended generation tasks like question answering and summarization, compared to overlap-based metrics. Appendix Table \ref{tab:nq_full} shows that exact-match accuracy can severely underestimate the quality of model outputs, while manual inspection confirms many responses deemed incorrect by traditional metrics are actually valid and possibly preferred due to natural phrasing or richer explanations. Despite the higher cost, we therefore recommend LLM-as-a-judge scoring as a valuable complement to traditional metrics and human evaluation for future work on LLM fine-tuning and benchmarking.

\subsection{Future Work}

The results and analysis in this work only partially capture the true effectiveness of the TAPR method. Both the datasets and prompts considered represent a narrow look at how people interact with LLMs in real-world scenarios. In practice, many user queries are more diverse, nuanced, and less well-formed than those found in common benchmarks. As such, small changes to questions or instructions on familiar datasets may yield limited improvements, while automatic prompt optimization could provide even greater benefits when adapting to users’ unique, everyday needs. This is partially supported by our finding that rewritten prompts tended to be clearer, more detailed, and more consistent with prompt engineering best practices than the baseline instructions, suggesting that our approach may have more potential not fully reflected by our evaluation metrics. In the future, a more robust test would be to deploy this method in real agentic workflows, where crafting effective prompts is a central challenge, and directly compare the outcomes achieved by human-written prompts versus automatically optimized prompts.

Our study was limited to smaller, non-reasoning models. Although scaling up to larger models is computationally expensive, it is also plausible that larger LLMs, with their broader knowledge and enhanced capabilities, could be more effective at both generating and benefiting from optimized prompts, particularly in the context of full-prompt rewriting where information loss is a concern. However, using advanced reasoning models that already employ chain-of-thought or step-by-step prompting may require specialized evaluation protocols and different prompt objectives to achieve maximum benefit.

Finally, one motivation for this approach was to help close the gap between small and large LLMs by maximizing the effectiveness of less capable models through better prompt engineering. If successful, this would enable cost savings by achieving similar downstream performance with less expensive models. However, our results show that the gains from prompt rewriting are not yet consistent enough to offset the additional training and inference costs. More detailed cost–benefit analyses, as well as further method refinements, will be valuable directions for future work.
\section{Conclusion}
\label{sec:conclusion}

In this work, we set out to address how to create a task-aware prompt rewriter model that effectively enhances LLM performance. Our findings demonstrate that a smaller model can be successfully trained through reinforcement learning to rewrite prompts that consistently improve downstream LLM outputs. Specifically, we showed that training with Group Relative Policy Optimization (GRPO) yields stable learning. Applying this trained TAPR model resulted in performance gains across various benchmarks. Notably, Phi-4-mini-instruct-based TAPR variants consistently outperformed baseline and base model prompts, delivering improved accuracy alongside qualitatively clearer, more detailed prompts adhering closely to established prompt engineering principles. Our ablation study revealed how the incorporation of LLM-as-a-judge evaluations, particularly the prompt quality reward, enhanced the training process, guiding the model towards clearer and more effective rewrites. We also observed LLM-as-a-judge evaluations to be more accurate than many traditional metrics.

Despite these promising results, our approach does have limitations, including variability in improvements across tasks and models, potential biases inherent to LLM-based reward evaluations, and increased computational demands. Future research should focus on refining the RL training and scaling to more diverse tasks and realistic user scenarios. 

Overall, this work demonstrates the viability of automated prompt rewriting as a practical and effective technique for unlocking the full potential of LLMs, reducing the reliance on manual prompt engineering.
\balance
\bibliography{references}

\newpage
\begin{acknowledgments}
  Thanks to the developers of ACM consolidated LaTeX styles
  \url{https://github.com/borisveytsman/acmart} and to the developers
  of Elsevier updated \LaTeX{} templates
  \url{https://www.ctan.org/tex-archive/macros/latex/contrib/els-cas-templates}.  
\end{acknowledgments}

\section*{Declaration on Generative AI}

 \noindent{\em Or (by using the activity taxonomy in ceur-ws.org/genai-tax.html):\newline}
During the preparation of this work, the author(s) used ChatGPT, Grammarly in order to: Grammar and spelling check, Paraphrase and reword, Improve writing style, Formatting assistance. After using this tool/service, the authors reviewed and edited the content as needed and take full responsibility for the publication’s content.


\appendix
\newpage
\begin{appendix}

\section{Prompts Used in Training and Evaluation}
\label{app:prompts}
\subsection{Meta-prompt for Instruction-Only TAPR Training}
\label{app:inst-only-meta}
\begin{lstlisting}
Rewrite the following instruction by rephrasing and/or adding specific requirements to ensure the best possible chance for another LLM to answer the query correctly. Use illustrative descriptions if needed. Don't add details that change the meaning of the instruction. Output your response as a JSON object with two keys: "explanation" and "final_rewritten_query". The "explanation" key should contain your reasoning for the changes you made, and the "final_rewritten_query" key should contain only the final rewritten instruction.

Here's a demonstrative example:
1. Original instruction: The old instruction
New Instruction: {"explanation": "Explanation of why this kind of rewriting was done", "final_rewritten_query": "The rewritten instruction."}

Now write the new instruction. Respond only with valid JSON. Do not write an introduction or summary.

Original instruction: "Answer the question"

New Instruction: {
\end{lstlisting}

\subsection{LLM-as-a-Judge Prompt for Question Answering}
\begin{lstlisting}
You are an AI assistant specialized in judging whether a given response correctly answers the user's query based on the reference answer(s).
Analyze [QUERY], the candidate [RESPONSE], and the list of acceptable [REFERENCE] answers, then decide if the response contains a correct answer.

Guidelines
1. Match at least one reference answer exactly or via an unambiguous synonym.
2. Ignore extra, irrelevant detail; focus on whether the core answer is present.
3. Output 1 for correct, 0 for incorrect.

Return a JSON object:
{
  "explanation": "...short justification...",
  "score": 1 | 0
}

Examples

[QUERY]     : Who wrote "Pride and Prejudice"?
[RESPONSE]  : Jane Austen wrote "Pride and Prejudice" in 1813.
[REFERENCE] : ["Jane Austen"]
{
  "explanation": "Mentions Jane Austen, which matches the reference exactly.",
  "score": 1
}

(3 more examples)

Now evaluate:
[QUERY]    : "{query}"
[RESPONSE] : "{final_answer}"
[REFERENCE]: "{target_answers}"
\end{lstlisting}

\section{Hyperparameters}
\label{sec:hyper}

We conduct the GRPO training using the TRL library \cite{huggingface2025trl}, applying the hyperparameter settings described here. To stabilize learning and prevent the policy from exploiting sequence‐length benefits, we adopt the \emph{Dr.\ GRPO} loss formulation and mask truncated completions so that only fully generated samples contribute to the gradient \cite{liu2025understanding}. Dropout is disabled to ensure that the reference model produces deterministic log‐probabilities for the KL divergence term, therefore reducing variance in the penalty. We generate four completions per prompt using sampling with maximum temperature \(T=1.0\), a top-\(k\) of 40, and nucleus sampling \(p=0.95\) to increase exploration. A warm-up ratio of 5 \% followed by cosine learning-rate decay prevents early optimization spikes, while a weight decay of 0.01 and global gradient clipping at 1.0 mitigate overfitting and exploding gradients. The learning rate of 1e-5 does not lead to a loss of capabilities while still allowing fairly quick convergence.

Each update comprises two GRPO iterations per batch and employs clipped-advantage bounds \(\epsilon_{\mathrm{low}}=0.2\) and \(\epsilon_{\mathrm{high}}=0.28\), maintaining learning momentum without catastrophic policy drift. Training in bfloat16 precision and 4-bit quantization with LoRA adapters enables us to conduct training with limited compute. These hyperparameter choices are informed by previous experimentation and literature mentioned in Section \ref{sec:ft}, striking a balance between stability and exploration to preserve the model’s text generation abilities while enabling effective prompt optimization. Further details can be seen in our codebase.

\section{Additional Experimental Results on Natural Questions}
\label{sec:nq_full}

In Table \ref{tab:nq_full}, we present results with another LLM in Qwen3-4B on the Natural Questions dataset. These results show how using Qwen3-4B as the base model did not prove to be as effective for our method as the other LLMs, signaling the importance of optimizing hyperparameters for each different setup. We also include F1, ROUGE, and exact-match accuracy scores to show how unreliable they can be compared to another LLM judging. 

\begin{table*}[ht]
\centering
\small
\setlength{\tabcolsep}{4pt}
\begin{adjustbox}{max width=\textwidth}
\begin{tabular}{lrrrr}
\toprule
\textbf{TAPR LLM} & \textbf{Accuracy (\%)} & \textbf{F1}  & \textbf{ROUGE}  & \textbf{Accuracy LLM judge(\%)} \\
\midrule
Baseline prompt         & 0.10 &  0.09 &  0.09 & 56.30 \\
\cdashline{1-5}[.5pt/2pt]
Phi-4-mini-instruct                                   & 0.00 &  0.02 &  0.02 & 48.80 \\
Phi-4-mini-instruct TAPR                   & 0.00 &  0.06 &  0.09 & \textbf{59.20} \\
Phi-4-mini-instruct TAPR + Selection                   & 0.00 &  0.07 &  0.08 & 57.30 \\
\cdashline{1-5}[.5pt/2pt]
Qwen3-4B                                              & 0.00 &  0.06 &  0.07 & \textbf{53.90} \\
Qwen3-4B TAPR                              & 0.00 &  0.07  &  0.08  & 44.40 \\
Qwen3-4B TAPR + Selection                                 & 0.00 &  0.08  &  0.10  & 47.35 \\
\cdashline{1-5}[.5pt/2pt]
LLaMA 3.2 3B-instruct                                  & 0.00 &  0.03  &  0.04  & 58.30 \\
LLaMA 3.2 3B-instruct TAPR                  & 0.00 &  0.06  &  0.07  & \textbf{62.20} \\
LLaMA 3.2 3B-instruct TAPR + Selection                     & 0.00 &  0.06  &  0.07  & 59.10 \\
\bottomrule
\end{tabular}
\end{adjustbox}
\caption{NQ dataset full results, including other metrics such as exact-match and including results from the Qwen models}\label{tab:nq_full}
\end{table*}

\section{Experimental Results - Full-Prompt Rewriting}
\label{sec:fpr}
Here, we introduce \textit{full-prompt rewriting}, where the TAPR LLM rewrites both the instruction and the specific input (e.g., the question). Therefore, both during training and evaluation, the model rewrites the default prompt for each sample. We observed that rewriting very long inputs can sometimes hurt performance; therefore, we limit full-prompt rewriting experiments to the two datasets with relatively short inputs
in Natural Questions and GSM8K. Furthermore, for full-prompt rewriting with the selection mechanism, we generate only 3 candidate rewrites per example to reduce computational overhead.

As seen in Table \ref{tab:fullprompt_combined}, we observe that full-prompt rewriting generally results in weaker performance compared to optimizing a single instruction for the entire dataset. This may be due to the loss of
important information during the rewriting process, even though we tried preserving the original prompt within the full input that the Task LLM receives. Nevertheless, sample-level full-prompt rewriting can offer greater robustness. If a single rewritten prompt is suboptimal, it only affects one sample, whereas a poor general prompt can negatively impact all predictions for a dataset or task. As it is a more complex rewriting task than rewriting one prompt per dataset, we hypothesize that a more capable LLM might be more suitable for this purpose.

\begin{table}[ht]
\centering
\begin{tabular}{l l c c}
\toprule
\textbf{TAPR LLM} & \textbf{Rewriting Variant} & \textbf{NQ $\uparrow$} & \textbf{GSM8K $\uparrow$} \\
\midrule
Baseline prompt & -- & \textbf{56.30} & 82.38 \\
\cdashline{1-4}[.5pt/2pt]
Phi-4-mini & Base Model & 49.70 & 83.00 \\
Phi-4-mini & TAPR & 53.95 & 77.60 \\
Phi-4-mini & TAPR + Selection & 53.16 & 78.20 \\
\cdashline{1-4}[.5pt/2pt]
Qwen3-4B & Base Model & 53.71 & 81.30 \\
Qwen3-4B & TAPR & 49.65 & 79.50 \\
Qwen3-4B & TAPR + Selection & 51.44 & \textbf{84.05} \\
\cdashline{1-4}[.5pt/2pt]
LLaMA-3.2-3B & Base Model & 53.15 & 79.00 \\
LLaMA-3.2-3B & TAPR & 51.10 & 79.60 \\
LLaMA-3.2-3B & TAPR + Selection & 49.35 & 77.70 \\
\bottomrule
\end{tabular}
\caption{Comparison of full-prompt rewriting accuracy across the NQ and GSM8K datasets. Values are accuracy percentages measured by GPT-4o-mini, reflecting answer correctness for each prompt variant.}
\label{tab:fullprompt_combined}
\end{table}

\section{Experimental Results - Alternative Task LLMs}
\label{sec:alter}

To assess the generality of our method, we evaluate TAPR performance with alternative Task LLMs, specifically Phi-4-mini-instruct and GPT-4o-mini. Results are presented in Table~\ref{tab:other}. They show how TAPR can work with other Task LLMs, most noticeably GSM8K performance with Phi-4-mini-instruct as the Task LLM goes from about 6\% accuracy to around 35\% with TAPR prompts. With a more capable Task LLM in GPT-4o-mini, the best results also come from the TAPR + Selection variant.

\begin{table}[h!]
  \centering
  \small
  \setlength{\tabcolsep}{6pt}
  \begin{tabularx}{\linewidth}{@{} l L L c c @{}}
    \toprule
    \textbf{TAPR LLM} & \textbf{Task LLM} & \textbf{Variant} & \textbf{NQ} $\uparrow$ & \textbf{GSM8K} $\uparrow$ \\
    \midrule
    Baseline            & Phi-4-mini & –                   & 31.10 & 18.90 \\
    \cdashline{1-5}[.5pt/2pt]
    Phi-4-mini         & Phi-4-mini & Base Model          & \textbf{33.80} & 6.10 \\
    Phi-4-mini         & Phi-4-mini & TAPR                & 31.70 & 34.90 \\
    Phi-4-mini         & Phi-4-mini & TAPR + Sel.    & 32.90 & \textbf{35.00} \\
    \cdashline{1-5}[.7pt/4pt]
    Baseline            & GPT-4o-mini & –                  & 63.16 & 77.56 \\
    \cdashline{1-5}[.5pt/2pt]
    LLaMA-3.2-3B       & GPT-4o-mini & Base Model          & 64.23 & 59.00 \\
    LLaMA-3.2-3B       & GPT-4o-mini & TAPR                & 64.00 & 86.40 \\
    LLaMA-3.2-3B       & GPT-4o-mini & TAPR + Sel.    & \textbf{64.50} & \textbf{88.40} \\
    \bottomrule
  \end{tabularx}
  \caption{Prompt rewriting performance with alternative Task LLMs (Accuracy \%). Results are shown for NQ and GSM8K.}
  \label{tab:other}
\end{table}

\section{Experimental Results - Generalization}
\label{sec:gen}
\begin{figure}[h!]
  \centering
  \begin{subfigure}
    \centering
    \includegraphics[width=\linewidth]{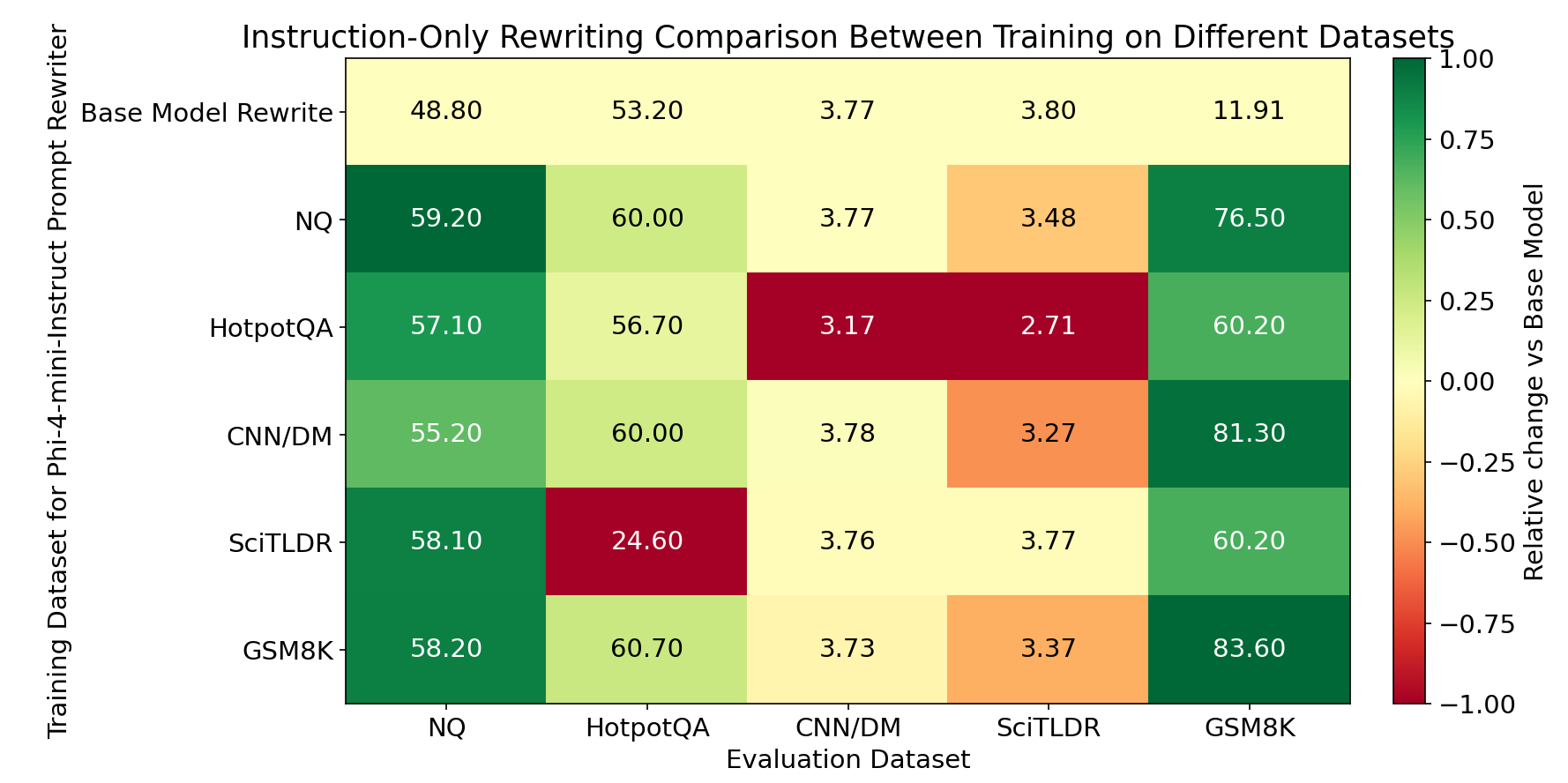}
    \caption{Cross‐dataset generalization of TAPR. Each cell reports the accuracy when Phi-4-mini-instruct is trained on the dataset that the row is named after, and evaluated on the dataset that the column is named after. Summarization scores remain on a 1–5 scale, while QA and GSM8K are percentages. Cell shading encodes change relative to the base prompt (top row), with green indicating higher accuracy and red indicating lower, and each column’s colormap is scaled independently.}
    \label{fig:heatmap}
  \end{subfigure}%
\end{figure}

To assess how task-aware our method is, we show results on how training on one dataset relates to performance on other datasets. Figure \ref{fig:heatmap} illustrate cross-dataset generalization through heatmaps, where each cell reports accuracy when a TAPR is trained on the row’s dataset and evaluated on the column’s dataset. We see that while training on the specific dataset mostly leads to the best performance, prompt rewriting ability does often transfer from one dataset to another, with training on NQ and GSM8K datasets leading to the most gains. In Table \ref{tab:multi_dataset} we report results from training on multiple datasets, and see that while it is less effective than single dataset training, it still leads to some gains on non-summarization datasets. 

\begin{table}[h!]
  \centering
  \resizebox{\columnwidth}{!}{%
    \begin{tabular}{@{} l l c c c c c @{}}
      \toprule
      \textbf{TAPR LLM} & \textbf{Variant} & \textbf{NQ} & \textbf{Hotpot} & \textbf{CNN/DM} & \textbf{Sci} & \textbf{GSM} \\
      \midrule
      Baseline & Phi-4-mini & 56.30 & 53.20 & 3.373 & 3.182 & 82.38 \\
      \cdashline{1-7}[.5pt/2pt]
      Phi-4-mini-instruct & Base Model & 48.80 & 53.20 & \textbf{3.782} & \textbf{3.802} & 11.91 \\
      Phi-4-mini-instruct & Multi TAPR & \textbf{57.90} & \textbf{56.30} & 3.362 & 3.068 & \textbf{82.70} \\
      \bottomrule
    \end{tabular}%
  }
  \caption{Performance of Phi-4-mini-instruct under TAPR training across multiple tasks. “Base Model” reports the untrained prompt performance. “Multi-dataset TAPR” corresponds to training on SciTLDR, GSM8K, and NQ jointly, with evaluation performed independently on each benchmark. Question answering and GSM8K results are reported as accuracy percentages, while CNN/DM and SciTLDR summarization scores are reported on a 1–5 scale. Best results within each column are shown in bold.}
  \label{tab:multi_dataset}
\end{table}

\end{appendix}

\newpage

\end{document}